\documentclass{article}

\usepackage[numbers,sort]{natbib}
     \usepackage[preprint]{neurips_2019}

\usepackage{amsmath}
\usepackage{textcomp}

\usepackage[utf8]{inputenc} 
\usepackage[T1]{fontenc}    
\usepackage{wrapfig}

\usepackage{hyperref}       
\usepackage{url}            
\usepackage{booktabs}       
\usepackage{amsfonts}       
\usepackage{nicefrac}       
\usepackage{microtype}      

\usepackage{graphicx}
\usepackage{algorithm}
\usepackage[noend]{algpseudocode}
\usepackage[colorinlistoftodos,prependcaption,textsize=tiny]{todonotes}

\graphicspath{{./}{./plots/}}
\begin{document}

\title{Semi-Supervised Natural Language Approach for Fine-Grained Classification of Medical Reports}

\author{
Neil Deshmukh \\
Moravian Academy | Lehigh University; Bethlehem, PA 15018 \\
\texttt{neil.nitin.de@gmail.com} \\\\
\and
\and Selin Gumustop\textsuperscript{1} \\
\texttt{sgumustop@partners.org}\\\\
\and Romane Gauriau\textsuperscript{1} \\
\texttt{romane.gauriau@mgh.harvard.edu} \\
\and Varun Buch\textsuperscript{1} \\
\texttt{varun.buch@mgh.harvard.edu} \\\\
\and Bradley Wright\textsuperscript{1} \\
\texttt{bwright9@partners.org} \\
\and Christopher Bridge\textsuperscript{1} \\
\texttt{cbridge@partners.org} \\
\and Ram Naidu\textsuperscript{1} \\
\texttt{rnaidu@partners.org} \\\\
\and Katherine Andriole\textsuperscript{1}\\
\texttt{kandriole@bwh.harvard.edu}
\and Bernardo Bizzo\textsuperscript{1} \\
\texttt{bbizzo@mgh.harvard.edu} \\\\\\\\
\textsuperscript{1}MGH \& BWH Center for Clinical Data Science; Boston, MA 02114
}
\maketitle
\begin{abstract}

Although machine learning has become a powerful tool to augment doctors in clinical analysis, the immense amount of labeled data that is necessary to train supervised learning approaches burdens each development task as time and resource intensive. The vast majority of dense clinical information is stored in written reports, detailing pertinent patient information. The challenge with utilizing natural language data for standard model development is due to the complex and unstructured nature of the modality. In this research, a model pipeline was developed to utilize an unsupervised approach to train an encoder-language model, a bidirectional recurrent neural network, to generate document encodings; which then can be used as features passed into a decoder-classifier model that requires magnitudes less labeled data than previous approaches to differentiate between fine-grained disease classes accurately. The language model was trained on unlabeled radiology reports from the Massachusetts General Hospital Radiology Department (n=218,159) and terminated with a loss of 1.62 and a word prediction accuracy of 62\%. The classification models were trained on three labeled datasets of head CT studies of reported patients, presenting large vessel occlusion (n=1403), acute ischemic strokes (n=331), and intracranial hemorrhage (n=4350), to identify a variety of different findings directly from the radiology report data; resulting in AUCs of 0.98, 0.95, and 0.99, respectively, for the large vessel occlusion, acute ischemic stroke, and intracranial hemorrhage datasets. The output encodings are able to be used in conjunction with imaging data, to create models that can process a multitude of different modalities. The ability to automatically extract relevant features from textual data allows for faster model development and integration of textual modality, overall, allowing clinical reports to become a more viable input for more encompassing and accurate deep learning models.

\end{abstract}

\section{Introduction}
Electronic Medical Record (EMR) data is the current standard system for the collection of medical information about a patient, and contains the vast majority of dense clinical information. The number of practices adopting an EMR system has been steadily increasing over the last few years (see Fig. 1), and has become a staple within healthcare facilities \cite{b1}.  There is an incredibly large amount of data within these systems, allowing for incredible insight with machine learning techniques, especially through textual data. Clinical reports from various sources, such as radiology exams, are commonly stored in the EMR systems, encompassing not only a comprehensive representation of the patients' status, but an accompanying analysis by the medical practitioner themselves, embedding an untapped clinical inference within these documents \cite{b2}. 

The cohort generation process for the development of datasets for machine learning applications requires the selection of appropriate data, based on specific criteria. Among several methods of such cohort selection for medical imaging data, existing radiology reports generated for clinical interpretation can be used to identify exams containing findings of interest. The radiology reports are commonly manually reviewed and annotated by experts to generate cohorts for medical imaging machine learning models, which is a time-consuming process that can be a bottleneck for development. 

Extracting this clinical information from within the vast amount of medical documents has been an immense challenge, due to the significant variations in medical terminology, and the complex, unstructured nature of textual data. Practitioners display a wide array of variability, which makes it difficult to utilize keyword searches or any form of supervised training, as the labeling process of such a large amount of data would be a monumental effort. Standard text encoding processes are unable to fully understand the contextual relationship found in medical reports \cite{b3}. 

In terms of the analysis for this initial research, acute strokes were the focus, due to their impact on the general populace. There are 795 thousand acute strokes in the US annually, and, taking more than 140 thousand lives per year, they are the third leading cause of death \cite{b4}.  Out of this immense number of strokes, 87\% are ischemic, in which blood flow to the brain is blocked \cite{b5}.  As such, evidence of hemorrhage contraindicates IA/IV treatment \cite{b6}, so it is relevant to future treatment options; 38.7\% of the subset of acute ischemic strokes presents large-vessel occlusion \cite{b7}.  Overall, acute ischemic stroke is a leading cause of long-term and preventable disability, all the while costing the US an estimated \$34 billion each year.

The increasing prevalence of EMR data through healthcare systems made it imperative to develop systems to automatically learn encodings of complex textual data. The purpose of this research was to develop an algorithm to automatically classify head CT radiology reports for specific stroke findings. 
\begin{figure}[htbp]
\centerline{\includegraphics[scale=0.25]{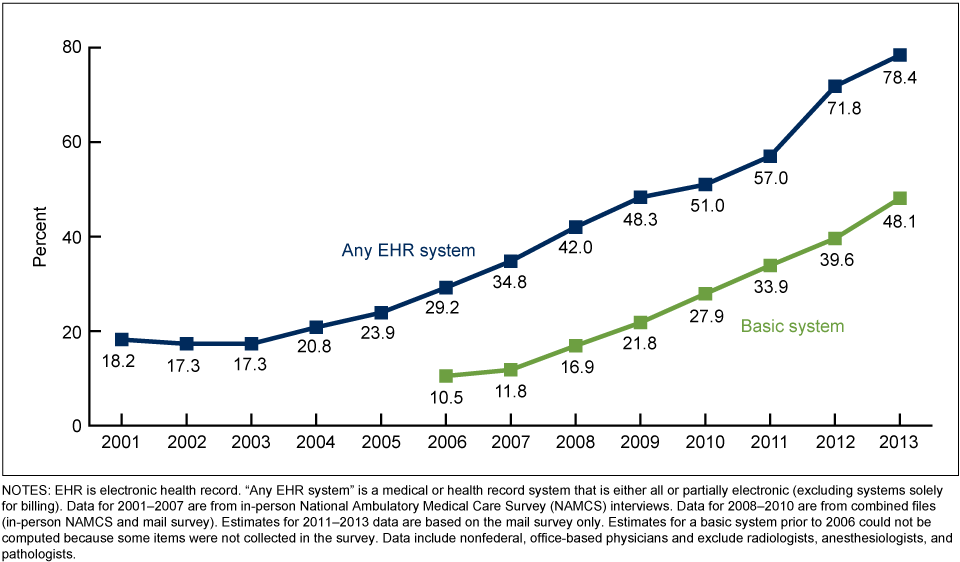}}
\caption{Graph of Practices with EHR Systems\cite{b1}}
\label{fig}
\end{figure}

\section{Methods}

\subsection{Database Preprocessing}
The entire text corpus utilized in the development of the language model was aggregated from the Massachusetts General Hospital (MGH) Department of Radiology archive. The dataset consisted of 218,159 radiology reports detailed by medical practitioners. The dataset was split into training, validation, and testing sets in a ratio of 85/15/5. 

 Words that showed up less than five times in the dataset were excluded from the vocabulary, and were usually typos in the reports or specific proper nouns. The dataset was also tokenized through the spaCy format to allow for better feature extraction by the model, utilizing tokens for uppercase characters and repetition. Each word input is set as a one-hot encoded vector as an input to the language model, corresponding to a vocabulary index and word embedding in the initial layer.
 
 It should be noted that there is a clear distinction between the general unlabeled dataset, which is utilized for the unsupervised encoder training, and the classification dataset, explained below, which contains labels and is utilized for the supervised decoder training. 

The labeled datasets were collected through hand-labeled reports by expert clinicians, who parsed through each report individually, and identified the specific labels necessary for classification. 

 The development set ratios were determined by the number of data points in the set. The labeled datasets for large vessel occlusion (n=1403), acute ischemic stroke (n=331), and intracranial hemorrhage (n=4350), had development set splits of 30/20/50, 30/10/60, and 10/20/70, respectively. The same text preprocessing for the language model occurs for the inputs in the classification set. The output for the acute ischemic stroke model is binary (presence or absence of acute ischemic stroke); meanwhile, the class outputs for the multilabel large vessel occlusion model are: M1 (first segment of the middle cerebral artery), M2 (second segment of the middle cerebral artery), distal ICA (internal carotid artery), and Basilar; and the class outputs of the multilabel intracranial hemorrhage model are: SDH (subdural hemorrhage), SAH (subarachnoid hemorrhage), IVH (intraventricular hemorrhage), IPH (intraparenchymal hemorrhage), and EDH (epidural hematoma).
 
\subsection{Data Exploration}
Even from preliminary analysis, the dataset of patient reports contained great variability in the expression of conditions. The amount of dense information within each report was difficult to parse, as doctors embedded many diagnoses and negatives within a single paragraph. 

Analyzing the most frequent n-grams within differing classes also did not identify a clear pattern, as a simple keyword search would be unable to contextualize the report as a whole, including the negation of particular conditions. The classes, even if simplified to the base positive and negative cases for a medical anomaly,  contained few defining characteristics indicative of a classification.

Shorter results tended to be a negative label, but, in some instances, the impression had descriptions of many different conditions, and with negation challenging to detect, it would be extremely difficult for the model to interpret without understanding context.

As part of the data exploration process, a scatter plot utilizing Term Frequency Inverse Document Frequency (TFIDF) was generated. The TFIDF value of a word is generally defined as:

\[tfidf( t, d, D ) = tf( t, d ) \times idf( t, D )\]

The term-frequency term refers to the number of instances a term is found in a document, while the $idf$ term, demonstrating its inverse proportionality to the number of documents it is found within, is more specifically defined as:

\[idf( t, D ) = log \frac{ |  D  | }{ 1 + | \{ d \in D : t \in d \} | }\]

As TFIDF vectorizers generate sparse vectors, Singular Value Decomposition (SVD) was used to reduce the dimensionality of the data. Unfortunately, as was identifiable by viewing the plot, the TFIDF clusters did not seem to be separable. The TFIDF encodings were explored further, utilizing the cosine similarity function to view similarity between documents, defined as:

\[cos( x,  y) = \frac { x \cdot y}{|| x|| \cdot || y||}\]

After many iterations of report exploration, it was determined that the TFIDF values of documents were not a helpful metric in the clustering of similar reports, as there seemed to be no visible pattern in the reports embedded near one another in the high dimensional space of the TFIDF encodings.

Through this preliminary data analysis, it was identified that it would be difficult for a model to determine relevant encodings and generate classifications without the ability not only to understand the general context, but also the specialized medical vocabulary found in the dataset. More specifically, due to the magnitudes greater unlabeled data available, a model that would be able to generate relevant encodings utilizing an unsupervised approach would allow for scalability and ease-of-use in downstream applications; this was the goal of the project.

\subsection{Machine Learning Approaches}

While the classification of the medical reports found in the dataset seemed too fine-grained a task to classify for standard ML approaches, they were still attempted with the binary occlusion dataset to obtain a preliminary measure of performance. TFIDF vectorization was utilized as the preliminary input into statistical ML models: focusing on Naive Bayes, logistic regression, and Support Vector Machine (SVM). While a variety of other ML approaches were utilized, including random forest and k-nearest-neighbors, they did not achieve any significant results compared to the three analyzed further. The TFIDF models all performed worse than random classification (0.5). 

A purely statistical model seemed unable to understand context in the impressions, specifically with variable descriptions of similar medical phenomena, as it is difficult to extract fine-grained context from frequency scores. To provide the models with embedding inputs that contained high dimensional structure, and some contextual embeddings, a Bi-directional Encoder Representations from Transformer (BERT) model was transfer trained on an unlabeled set of reports (n=218,159). The large, cased BERT model subtype with embeddings generated from the concatenation of the final four layers performed optimally in testing. The encodings generated by the custom BERT model were used to vectorize the text as an input for the standard machine learning models described earlier. 

\begin{figure}[htbp]
\centerline{\includegraphics[scale=0.15]{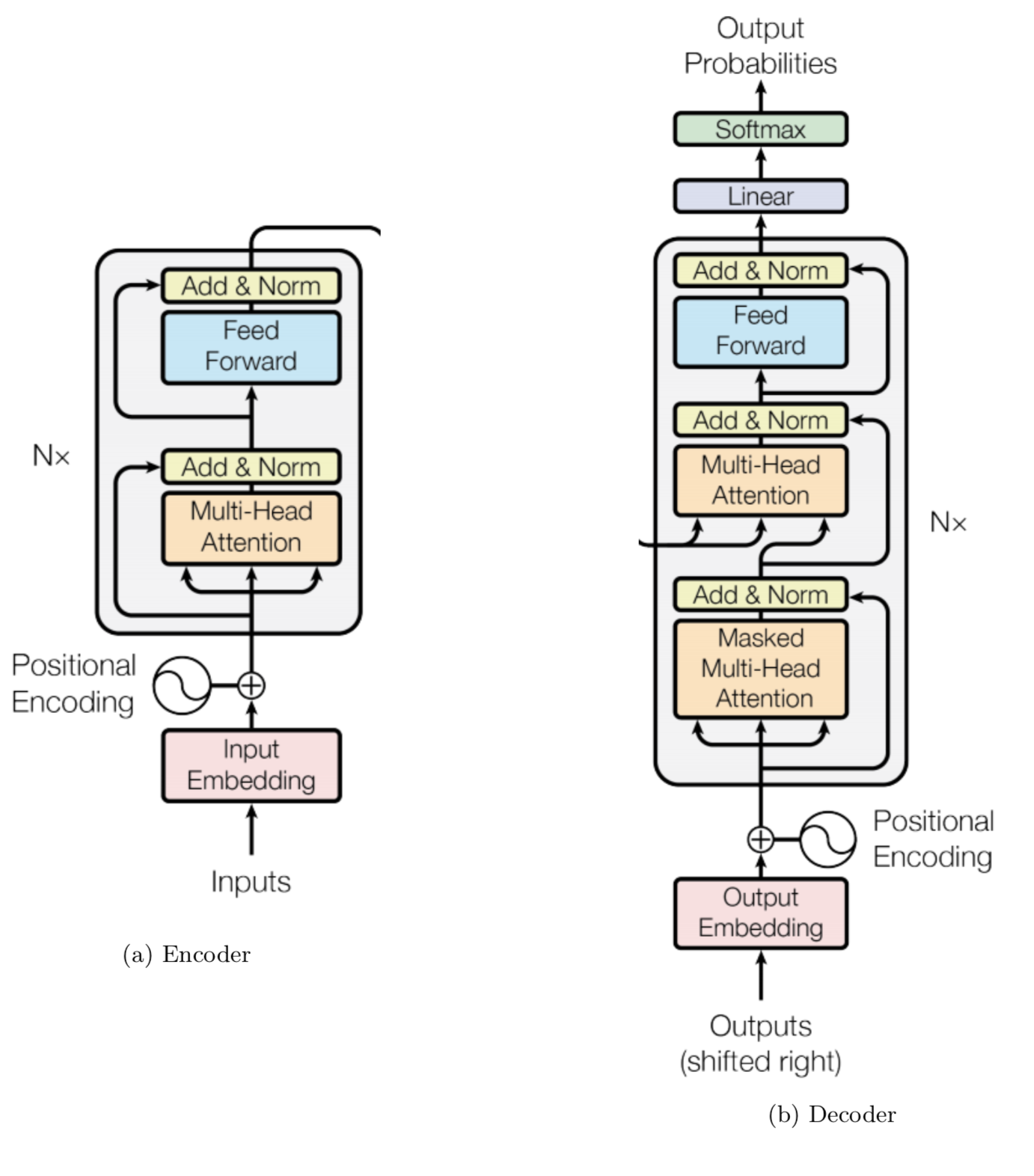}}
\caption{Structure of BERT Model\cite{b8}}
\label{fig}
\end{figure}

While the models did marginally better with BERT-embeddings, both the inability to pre-train the BERT model due to computational limitations, and the impairment to create fully context-based embeddings, due to the mask-based training structure, makes it difficult for fine-grained classification tasks.

In viewing that there seemed to be some context and language understanding within the BERT embeddings, a more complex fully connected, neural network (NN) was tested as the classifier, compared to the previous statistical approaches. The hope was that the NN would be able to identify more fine-grained features than a purely unsupervised network. The best architecture was one that was four layers of 512, 256, 128, to output, while L2 regularization and dropout were both used. The NN classifiers' performance was significantly higher compared to other previous approaches.

\begin{table}[htbp]
\caption{ML Model Results}
\begin{center}
\begin{tabular}{|c|c|c|}
\hline
 Model & F1-Score & AUC \\[0.5ex] 
 \hline\hline
  Naive Bayes (TFIDF) & 0.28 & 0.37\\ 
\hline
  Logistic (TFIDF) & 0.35 & 0.43 \\ 
  \hline
  SVM (TFIDF)& 0.33 & 0.42 \\ 
  \hline
  Neural Network (TFIDF) & 0.53 & 0.46 \\ 
 \hline
   Naive Bayes (BERT-Embedding) & 0.39 & 0.46\\ 
\hline
  Logistic (BERT-Embedding) & 0.41 & 0.68 \\ 
  \hline
  SVM (BERT-Embedding)& 0.44 & 0.39 \\ 
 \hline
    Neural Network (BERT-Embedding) & 0.61 & 0.72\\
 \hline
 \hline
\end{tabular}
\label{tab1}
\end{center}
\end{table}

\subsection{Development of Unsupervised Language Model}

While the NN improved the metrics compared to previous approaches, they were still not optimal for cohort generation or report classification.

As mentioned, the main challenge in processing medical reports is the variability in the description of diseases, making it difficult to classify with a small sample set. Techniques that are currently widely used, identify specific keywords through an iterative process, which is time-consuming and is not capable of accounting for the variability and complexity found in medical reports. Approaches that utilize models that do not understand contextual relations in the text are not capable of easily generalizing to different datasets. There was a need to optimize the cohort generation process to reduce the resources and time necessary for development \cite{b9}.

For these reasons, an unsupervised approach was utilized, simplifying the learning process by allowing the model to identify encodings, medical terminology, and context automatically, without any external work for a wide array of records. This allows for great scalability, simply by retraining the model on a larger corpus, were one to become available. An encoder model with the ability to generate single-size encodings is a powerful tool to create a high-dimensional representation of reports that are useful in a variety of downstream tasks.

The model architecture is a Recurrent Neural Network (RNN), which is specifically well suited to process textual data, due to their self-inputting nature, allowing for the detection of specific patterns in varying intervals of data. A many-to-one approach was utilized in this model, as a sequence was the input, and a word prediction classification was outputted. Both Gated Recurrent Units (GRUs) and Long Short Term Memory (LSTM) units were tested, and the LSTM networks resulted in a higher accuracy (see Fig. 3). The model is bidirectional, allowing for better context analysis from intervals of data. 

The Pytorch Machine Learning Library \cite{b10}, was used to develop, train, test, and optimize the neural network (NN). The RNN has a structure containing five layers and 31,683,132 trainable parameters. The model's preliminary embedding layer contains 28,362 parameters.

This model was determined after a period of research, as it allows for entirely unsupervised natural language understanding, specifically for medical reports. Processing each next word as the 'labeled' dataset transforms the unsupervised approach of language modeling into one that has corresponding word labels that allow the model to learn encodings for the report document into a lower-dimensional representation that simplifies the classification for the densely connected network. Through the process of self-identifying features, the model is capable of learning the context of medical terminology and developing encodings of the model that are relevant for downstream machine learning approaches.

\begin{figure}[htbp]
\centerline{\includegraphics[scale=0.2]{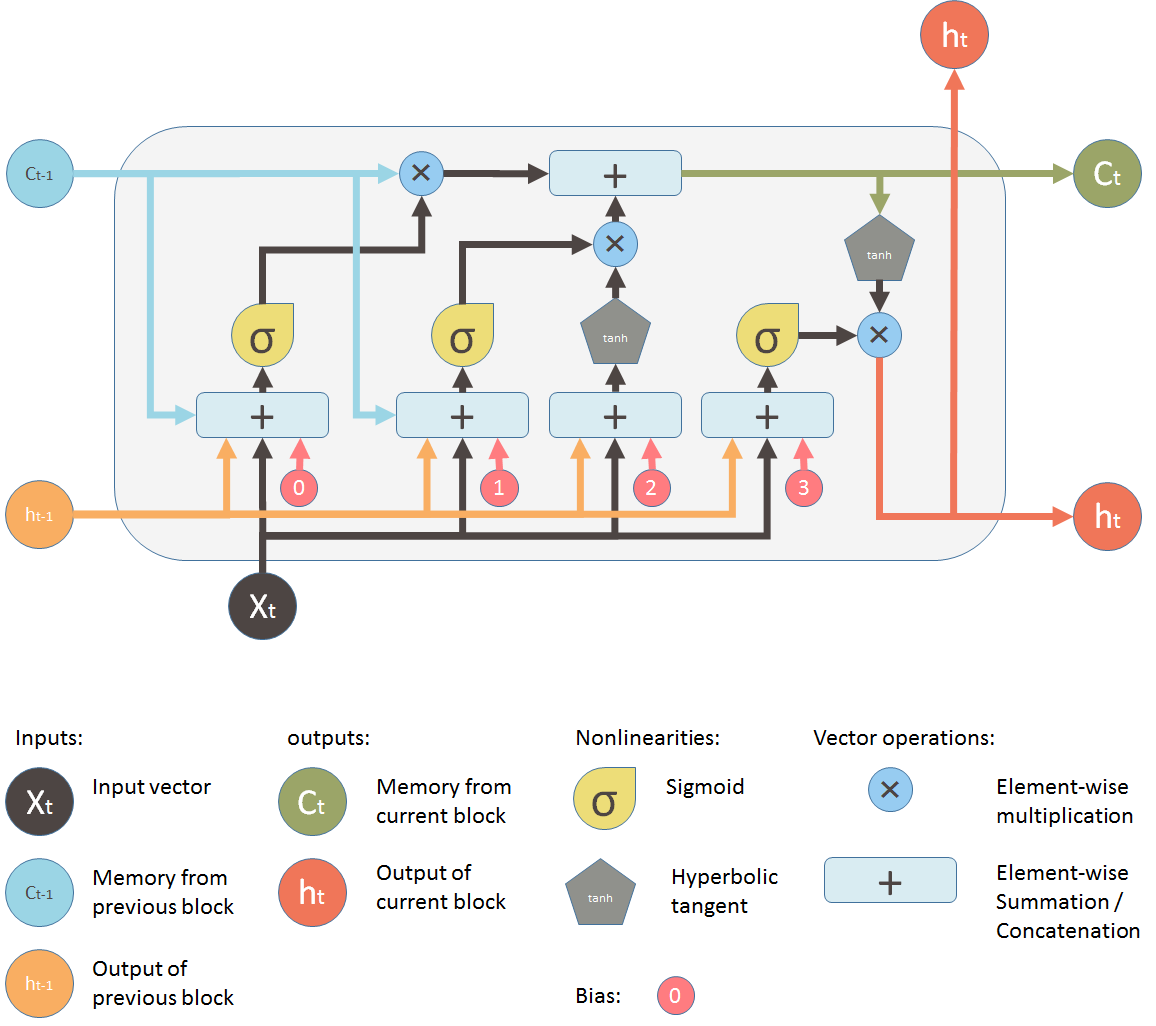}}
\caption{Structure of LSTM Unit \cite{b11}}
\label{fig}
\end{figure}

\subsection{Development of Classifier Model}

In order to process the encodings from the language model, the classifier inputs are the activations of the model from the layer before the final word classification layer, with an encoding size of 400. 

Due to the nature of the classifier head of the model, the RNN encoder can remain frozen while the classifier can be modified to accommodate any variety of inputs. The occlusion and hemorrhage classification models required a multilabel approach, with sigmoid activations on each of the corresponding output nodes to classes, as a patient could express multiple types of the conditions, while the stroke model was a binary classifier. Particularly, the models were trained with cross-entropy loss.

Both demonstrate the versatility of the language model, as if there are any clinical text machine learning analyses necessary, this model is capable of generating single-length representations of report documents that are high dimensional representations that allow for more efficient training pipelines.

\subsection{Training of Neural Networks}
Once the NNs had been developed, they were trained solely on raw inputs, taking in the tokenized text without any handcrafted features. Utilizing a deep learning approach increases the usability of this model by allowing for this unsupervised approach. The RNN was trained to optimize for smaller matrix operations, to allow them to run inference on large datasets in a reasonable amount of time on a standard desktop system. 

To emphasize, the RNN encoder model was trained on the unlabeled dataset (n=218,159) of radiology reports, and frozen when the optimal loss was achieved. The classifier models were then trained on top of the encoder, on the much smaller labeled datasets, for 500 epochs, after which the encoder was unfrozen to update parameters incrementally. Overall, it was trained over the course of 1500 epochs with cross-entropy loss and learning rate of 0.0003.

All of the previously mentioned training was conducted on a research computer, utilizing 32 GBs of GPUs, 16 vCPUs and, 20 GBs of RAM.

\section{Results}
The language model RNN was evaluated with the testing loss, as accuracy is not an intuitive metric for optimization. The testing set (n=10,908) comprised 5\% of the entire unlabeled report dataset. The final result was a loss of 1.62, with a vocabulary and embedding size of 18,362. 
\begin{table}[htbp]
\caption{Semi-Supervised Model Results}
\begin{center}
\begin{tabular}{|c|c|c|}
\hline
 Dataset & F1-Score & AUC \\[0.5ex] 
 \hline\hline
  Occlusion  & 0.96 & 0.98\\ 
\hline
  Stroke  & 0.92 & 0.95\\ 
  \hline
  Hemorrhage  & 0.98 & 0.99\\ 
 \hline
 \hline
\end{tabular}
\label{tab1}
\end{center}
\end{table}

The encodings from the second-to-last layer of the model were transformed to 3D space through Singular Vector Decomposition (SVD), an algorithm that identifies orthogonal values to transform high-dimensional representations by simplifying the data. (see Fig. 4)

\begin{figure}[htbp]
\centerline{\includegraphics[scale=0.4]{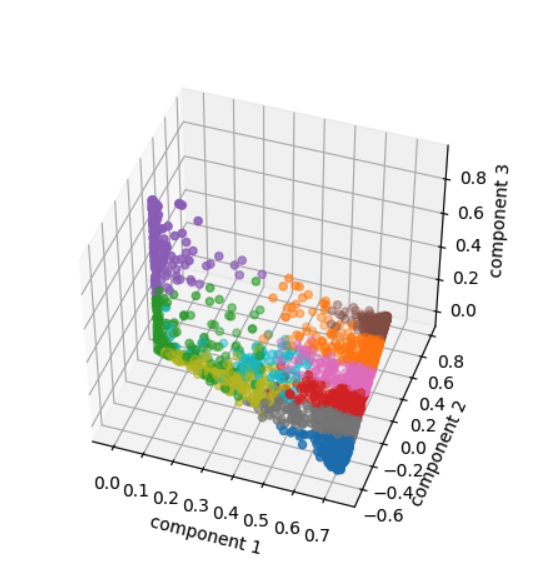}}
\caption{Plot of Encodings from RNN through PCA}
\label{fig}
\end{figure}

Area Under the Receiver Operator Characteristic (AUROC) curve, which plots the true positive rate against the false positive rate, was the metric that was optimized to allow for thorough analysis, while F1-Score was also calculated in order to analyze the performance of the model further. In terms of both, a micro-averaging strategy was utilized to determine a singular metric across all classes. With a large number of classes, F1-Score allows a metric that optimizes to make sure that the algorithm identifies the classes correctly without many false predictions, specifically when the dataset is imbalanced. The models achieved AUCs of 0.98, 0.95, and 0.99 on the datasets on occlusions, stroke, and hemorrhages, respectively.

\begin{figure}[htbp]
\centerline{\includegraphics[scale=0.4]{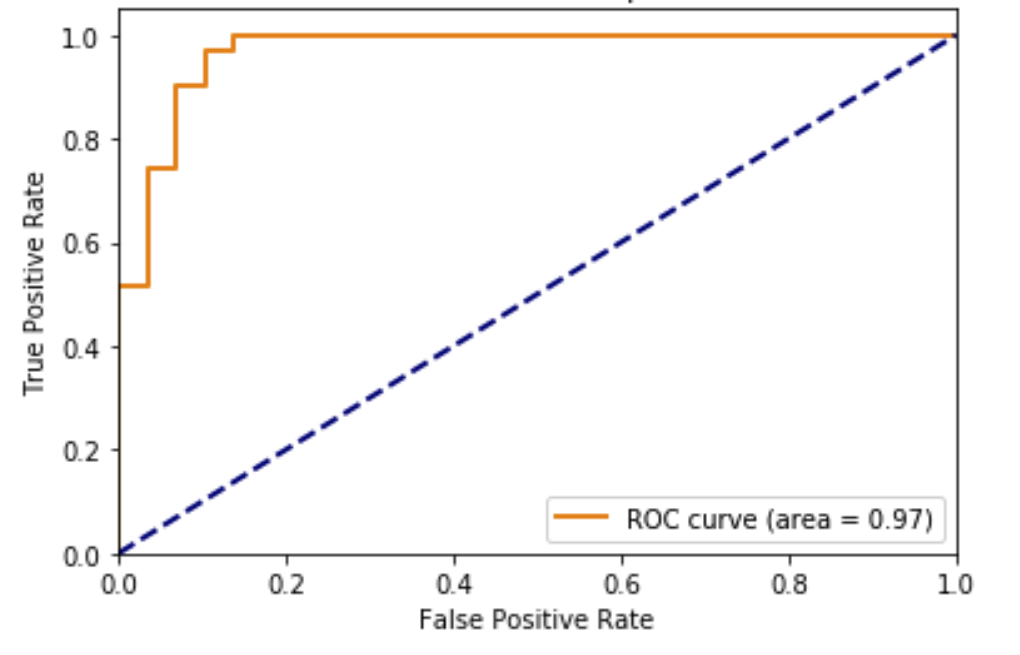}}
\caption{Receiver Operator Characteristic Curve for Occlusion Model}
\label{fig}
\end{figure}

\begin{figure}[htbp]
\centerline{\includegraphics[scale=0.4]{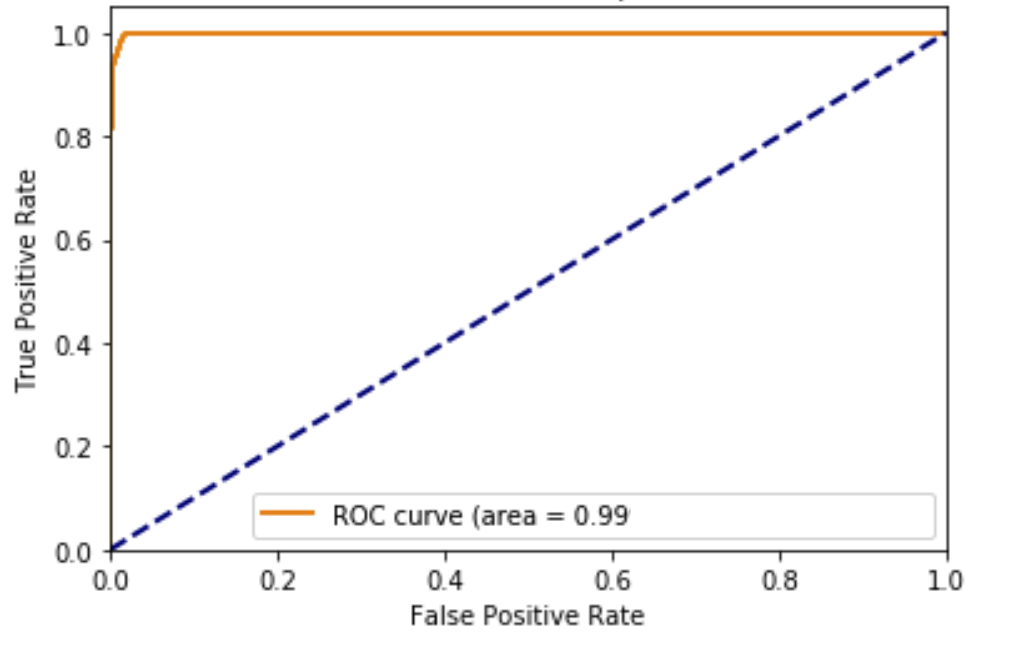}}
\caption{Receiver Operator Characteristic Curve for Hemorrhage Model}
\label{fig}
\end{figure}

\section{Conclusion}
The semi-supervised models and process outlined in this research achieved all the goals outlined and far exceed the current standard for medical record classification. By conducting completely automatic encoding of medical documents, across a wide array of subfields, this language model allows for the development of extremely well-performing models, simply utilizing magnitudes less labeled data. This scalable method results in a more efficient development pipeline for data curation, requiring a few samples to identify patients in a particular subset, classifying medical reports even if similar, through fine-grained classification. The language model, once trained, has a numerous amount of downstream uses, from text clustering, to cohort generation, to the generation of encodings as another modality input into standard ML algorithms. Overall, this model allows for the automated generation of encodings that contain relevant clinical information from medical report data, which has the potential to leverage an enormous set of patient data in the medical system, and lead to the development of more comprehensive models with a multitude of modalities.

\bibliography{bibliography}

\end{document}